\documentclass{article}
\usepackage{tcolorbox}
\usepackage{amsmath}
\usepackage{amssymb}
\usepackage{xcolor}      
\usepackage{graphicx}      
\usepackage{colortbl}    
\usepackage{array}       
\usepackage{booktabs,multirow}
\usepackage[table]{xcolor}
\usepackage{caption}   

\definecolor{hessiancolor}{RGB}{204, 57, 42}    
\definecolor{ggncolor}{RGB}{79, 155, 143}       
\definecolor{bhessiancolor}{RGB}{44, 97, 194}   
\definecolor{bggncolor}{RGB}{217, 116, 89}      
\definecolor{ekfaccolor}{RGB}{228, 197, 119}    
\definecolor{kfaccolor}{RGB}{155, 106, 145}     

\definecolor{hessianbg}{RGB}{245, 235, 234}     
\definecolor{ggnbg}{RGB}{235, 243, 242}         
\definecolor{bggnbg}{RGB}{245, 238, 235}        
\definecolor{ekfacbg}{RGB}{250, 247, 238}       
\definecolor{kfacbg}{RGB}{243, 238, 242}        

\definecolor{sectionbg}{RGB}{230, 230, 240}     

\PassOptionsToPackage{numbers, compress}{natbib}
 \usepackage[dblblindworkshop, final]{neurips_2025}

\usepackage[utf8]{inputenc} 
\usepackage[T1]{fontenc}    
\usepackage[%
  colorlinks = true,
  citecolor  = hessiancolor,
  linkcolor  = hessiancolor,
  urlcolor   = blue,
  unicode,
  linktocpage,
]{hyperref}        
\usepackage{url}            
\usepackage{booktabs}       
\usepackage{amsfonts}       
\usepackage{nicefrac}       
\usepackage{microtype}      
\usepackage{xcolor}         

\title{Multi-scale Autoregressive Models are Laplacian, Discrete, and Latent Diffusion Models in Disguise}
\workshoptitle{SPIGM workshop @ NeurIPS 2025}

\author{
  Steve Hong\\
  University of Cambridge\\
  \texttt{mdh58@cam.ac.uk}
  \And
  Samuel Belkadi\\
  University of Cambridge\\
  \texttt{bk2764@cam.ac.uk}
}

\begin{document}

\maketitle

\begin{abstract}
We reinterpret Visual Autoregressive (VAR) models as iterative refinement models to identify which design choices drive their quality–efficiency trade-off. Instead of treating VAR only as next-scale autoregression, we formalise it as a deterministic forward process that builds a Laplacian-style latent pyramid, together with a learned backward process that reconstructs samples in a small number of coarse-to-fine steps. This formulation makes the link to denoising diffusion explicit and highlights three modelling choices that may underlie VAR’s efficiency and sample quality: refinement in a learned latent space, discrete prediction over code indices, and decomposition by spatial frequency. We support this view with controlled experiments that isolate the contribution of each factor to quality and speed. We also discuss how the same framework can be adapted to permutation-invariant graph generation and probabilistic medium-range weather forecasting, and how it provides practical points of contact with diffusion methods while preserving few-step, scale-parallel generation.

\end{abstract}

\section{Introduction}

Generative modelling is often framed through two complementary paradigms, each with clear strengths and limitations \citep{yang2023diffusion}. Next-token autoregressive (AR) models are effective for discrete sequence generation (e.g., text) \citep{vaswani2017attention}, but they tend to struggle in high-fidelity image synthesis because imposed causal orderings, such as raster scan, disrupt natural spatial structure \citep{oord2016conditionalimagegenerationpixelcnn, tian2024visual}. Diffusion models (DDPM), by contrast, perform strongly on continuous data such as images and audio \citep{ho2020denoising, dhariwal2021diffusionmodelsbeatgans}. Their main practical limitation is computational: high-quality sampling often requires hundreds to thousands of refinement steps, and discrete outputs require additional modelling machinery \citep{ho2020denoising, austin2021structured}.

Against this backdrop, Visual Autoregressive (VAR) models \citep{tian2024visual} were introduced as a coarse-to-fine alternative that narrows the fidelity gap between diffusion and AR. VAR predicts each higher-resolution scale conditioned on lower-resolution scales, while generating all tokens within a scale in parallel rather than token by token. This scale-parallel strategy yields competitive image quality with substantially fewer refinement iterations, which improves inference speed in practice.

This naturally raises the question of what exactly drives those gains. VAR attributes much of its performance to next-scale prediction \citep{tian2024visual}, yet coarse-to-fine refinement itself is not new. Classical Laplacian pyramids already use this principle \citep{burt1983multiresolution}, and several diffusion variants also incorporate multi-scale structure \citep{ho2021cascadeddiffusionmodelshigh, saharia2021imagesuperresolutioniterativerefinement, fan2023frido, atzmon2024edify}. The key issue, then, is whether scale refinement alone explains VAR's combined fidelity and speed. We therefore focus on a more explicit attribution analysis.

\begin{figure*}[t]
  \centering
  \includegraphics[width=\textwidth]{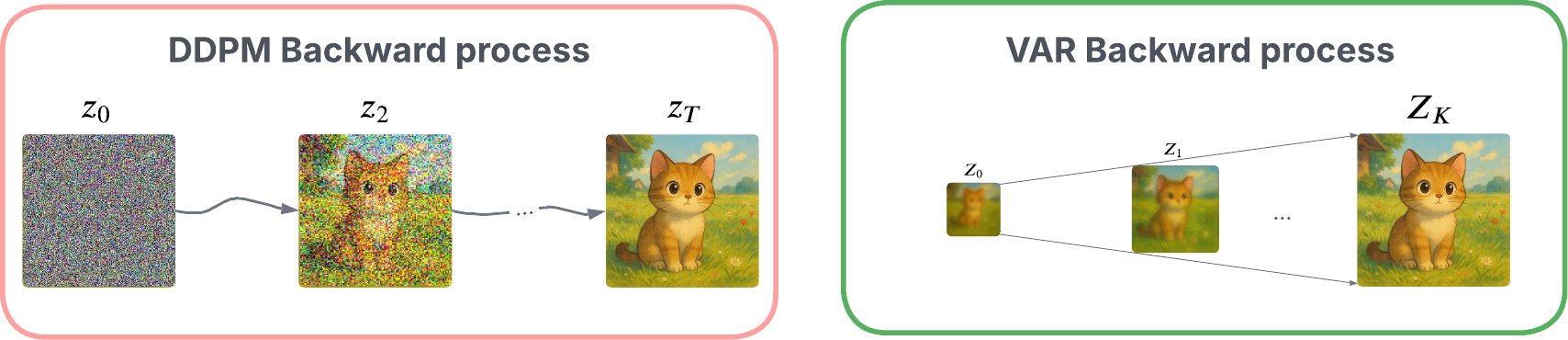}
    \caption{Iterative refinement in two paradigms. VAR refines by predicting next-scale discrete codes across a small number of scales; each step adds a quantised residual that captures band-limited detail (with finer scales carrying higher-frequency content). DDPM refines by denoising continuous latents over many timesteps; at each step, the model estimates and removes an approximately Gaussian noise component.}
  \label{fig:backward}
\end{figure*}

A related motivation comes from current research trajectories. Recent work has progressed along two largely separate tracks: incremental improvements to VAR and broader acceleration methods for diffusion. On the VAR side, studies mainly target engineering bottlenecks, including collaborative decoding for KV-cache and long-sequence overhead \citep{chen2024collaborativedecodingmakesvisual}, first-order Markov localisation of scale refinement for efficiency \citep{zhang2025mvarvisualautoregressivemodeling}, and infinite-vocabulary classification with bitwise self-correction to ease codebook bottlenecks \citep{han2025infinity}. On the diffusion side, effort has focused on reducing either step count or per-step cost through consistency objectives \citep{song2023consistency}, progressive distillation \citep{salimans2022progressive}, and fast ODE solvers \citep{lu2022dpm}, often with latent-consistency adapters. A unified iterative-refinement view can connect these lines of work and clarify where methods may transfer between families.

\paragraph{Core contributions.}
In this work, we reframe VAR as an iterative refinement model rather than a purely multi-scale autoregressive model. This perspective yields three concrete contributions:
\begin{enumerate}
\item A formal connection between VAR and denoising diffusion via deterministic forward decomposition and learned backward refinement in latent space.
\item Controlled ablations that isolate three factors behind VAR's efficiency-quality trade-off: latent representation, discrete targets, and coarse-to-fine frequency partitioning.
\item Design implications for non-image generation settings, with concrete hypotheses for graph generation and probabilistic weather forecasting.
\end{enumerate}

\section{Background}

\subsection{Visual Autoregressive Models}

\textbf{VQ-VAE tokeniser.} VAR uses a vector-quantised autoencoder that turns an input image $x \in \mathbb{R}^{H \times W \times 3}$ into a grid of discrete codes. An encoder $E_\psi$ produces a continuous feature map $z_e = E_\psi(x)$. Each vector in $z_e$ is quantised to its nearest codebook embedding $e_i$ from a learnable codebook $\mathcal{E}=\{e_i\}_{i=1}^{V}$, yielding the quantised map $z_q$ that is fed to a decoder $D_\phi$ to reconstruct $\hat x = D_\phi(z_q)$. Training uses the VQ-VAE objective with a reconstruction term and the codebook/commitment terms:
\begin{equation}
\mathcal{L}_{\mathrm{VQ}} \;=\; \underbrace{\|x-\hat x\|_2^2}_{\text{reconstruction}}
\;+\; \underbrace{\|\,\mathrm{sg}[z_e]-e\,\|_2^2}_{\text{codebook}}
\;+\; \beta\,\underbrace{\|\,z_e-\mathrm{sg}[e]\,\|_2^2}_{\text{commitment}} .
\end{equation}
where $\mathrm{sg}[\cdot]$ denotes the stop-gradient operator. This loss causes the encoder to commit to discrete codes, while the codebook tracks encoder outputs via vector quantisation, providing discrete token grids suitable for autoregression. VAR replaces a single-scale grid with a stack of token maps $(r^{(1)},\dots,r^{(S)})$ at increasing resolutions produced by a multi-scale VQ tokeniser.

\textbf{Next-scale autoregression.} Let $r^{(k)}$ denote the token map at scale $k$. VAR models the joint as
\begin{equation}
p(r^{(1)},\ldots,r^{(S)})\;=\;\prod_{k=1}^{S} p\!\left(r^{(k)} \mid r^{(<k)}\right),
\end{equation}
and predicts all tokens within a scale in parallel, using attention that is conditioned only on the prefix lower-resolution maps $r^{(\le k-1)}$ (block-causal across scales). This shift from next-token to next-scale prediction preserves 2D locality and reduces generation complexity relative to raster AR, as each refinement step operates on a 2D grid at a single scale.

It is important to note that the tokeniser and generator operate in \emph{residual} scales: at level $k$ the encoder forms a residual token map by subtracting the upsampled contribution of coarser scales ($x\!\leftarrow\! x-\phi_k(z_k)$), and at generation time the model \emph{predicts residuals} and reconstructs by upsample-and-add across all scales ($\hat x\!\leftarrow\!\hat x+\phi_k(z_k)$). \citep{tian2024visual}

\subsection{Denoising Diffusion Probabilistic Models}

DDPMs \citep{ho2020denoising} define a fixed forward (noising) Markov chain
\begin{equation}
q(x_t \mid x_{t-1})=\mathcal{N}\!\big(\sqrt{1-\beta_t}\,x_{t-1},\,\beta_t I\big),\quad t=1,\dots,T,
\end{equation}
which admits a closed-form marginal
\begin{equation}
q(x_t \mid x_0)=\mathcal{N}\!\big(\sqrt{\bar\alpha_t}\,x_0,\,(1-\bar\alpha_t)I\big),\quad \bar\alpha_t=\textstyle\prod_{s=1}^{t}\alpha_s,\;\alpha_t=1-\beta_t .
\end{equation}
The reverse chain is learned as Gaussian conditionals
\begin{equation}
p_\theta(x_{t-1} \mid x_t)=\mathcal{N}\!\big(\mu_\theta(x_t,t),\,\Sigma_\theta(t)\big),
\end{equation}
by optimizing a variational bound that decomposes into tractable KL terms between Gaussians (thanks to the closed-form forward posterior $q(x_{t-1} \mid x_t,x_0)$). In practice, parameterising $\mu_\theta$ via $\varepsilon_\theta$ (predicting the added noise) yields the simplified training step
\begin{equation}
\min_\theta\;\mathbb{E}_{t,x_0,\varepsilon}\,\big\|\,\varepsilon-\varepsilon_\theta\!\big(\sqrt{\bar\alpha_t}\,x_0+\sqrt{1-\bar\alpha_t}\,\varepsilon,\;t\big)\big\|_2^2,
\end{equation}
and sampling iterates $x_T \sim \mathcal{N}(0,I) \rightarrow x_0$ using the learned reverse conditionals.

There are two important remarks about DDPM:
\textbf{(i)} Closed-form marginals and posteriors make training low-variance and fully supervised at each noise level (the KL decomposition compares Gaussians in closed form).
\textbf{(ii)} Many-step generation: quality improves with the number of reverse steps $T$, but naïve DDPM sampling is slow; learning (or interpolating) reverse variances enables strong quality with far fewer steps (tens instead of hundreds), materially improving practicality.

\subsection{Laplacian pyramid}

The Laplacian pyramid \citep{burt1983multiresolution} represents an image as a low-pass residual plus a stack of band-pass detail layers arranged from coarse to fine. Let $G_0 = x$ be the original image and let $\text{down}(\cdot)$ denote low-pass filtering followed by subsampling. The Gaussian pyramid is built by
\begin{equation}
G_{k+1} \;=\; \text{down}(G_k), \qquad k=0,\dots,S-1.
\end{equation}
Define $\text{up}(\cdot)$ as expand (upsample) followed by low-pass filtering that is matched to the analysis filter. The Laplacian (detail) levels are then
\begin{equation}
L_k \;=\; G_k \;-\; \text{up}(G_{k+1}), \qquad k=0,\dots,S-1,
\end{equation}
together with the coarsest residual $G_S$. Perfect reconstruction follows by upsample-and-add:
\begin{equation}
\widehat{G}_S \;=\; G_S,\qquad 
\widehat{G}_k \;=\; L_k \;+\; \text{up}(\widehat{G}_{k+1})\ \ \text{for }k=S-1,\dots,0,
\end{equation}
yielding $\widehat{x}=\widehat{G}_0=x$ when the synthesis operator $\text{up}(\cdot)$ is the adjoint (or appropriately matched) to the analysis operator in $\text{down}(\cdot)$. Each $L_k$ behaves as a spatially localised band-pass component, while $G_S$ captures the low frequencies.

\section{Multi-scale Autoregressive Models as Iterative Refinement Models}

\subsection{Iterative refinement}

We view \emph{iterative refinement} as generating by a short sequence of learned updates on a state $x$. From an easy initialiser $x_0\sim\pi_0$, apply step-specific maps to approach the target:
\begin{equation}
x_{t+1} = \Phi_{\theta}^{(t)}(x_t, c), \qquad t=0,\dots,T-1,\quad \hat x = x_T .
\end{equation}
Each step is trained to make a local improvement, e.g. by decreasing a task energy or satisfying a consistency relation:
\begin{equation}
E(x_{t+1};c) \le E(x_t;c)-\delta_t \quad\text{or}\quad \mathcal{C}(x_{t+1},x_t,c)\approx 0 .
\end{equation}
A coarse-to-fine schedule arises by constraining $\Phi_{\theta}^{(t)}$, or the representation, such that early steps set global structure while later steps add detail; this applies to discrete or continuous $x$; in pixel, latent, or other spaces.

\begin{figure*}[t]
  \centering
  \includegraphics[width=\textwidth]{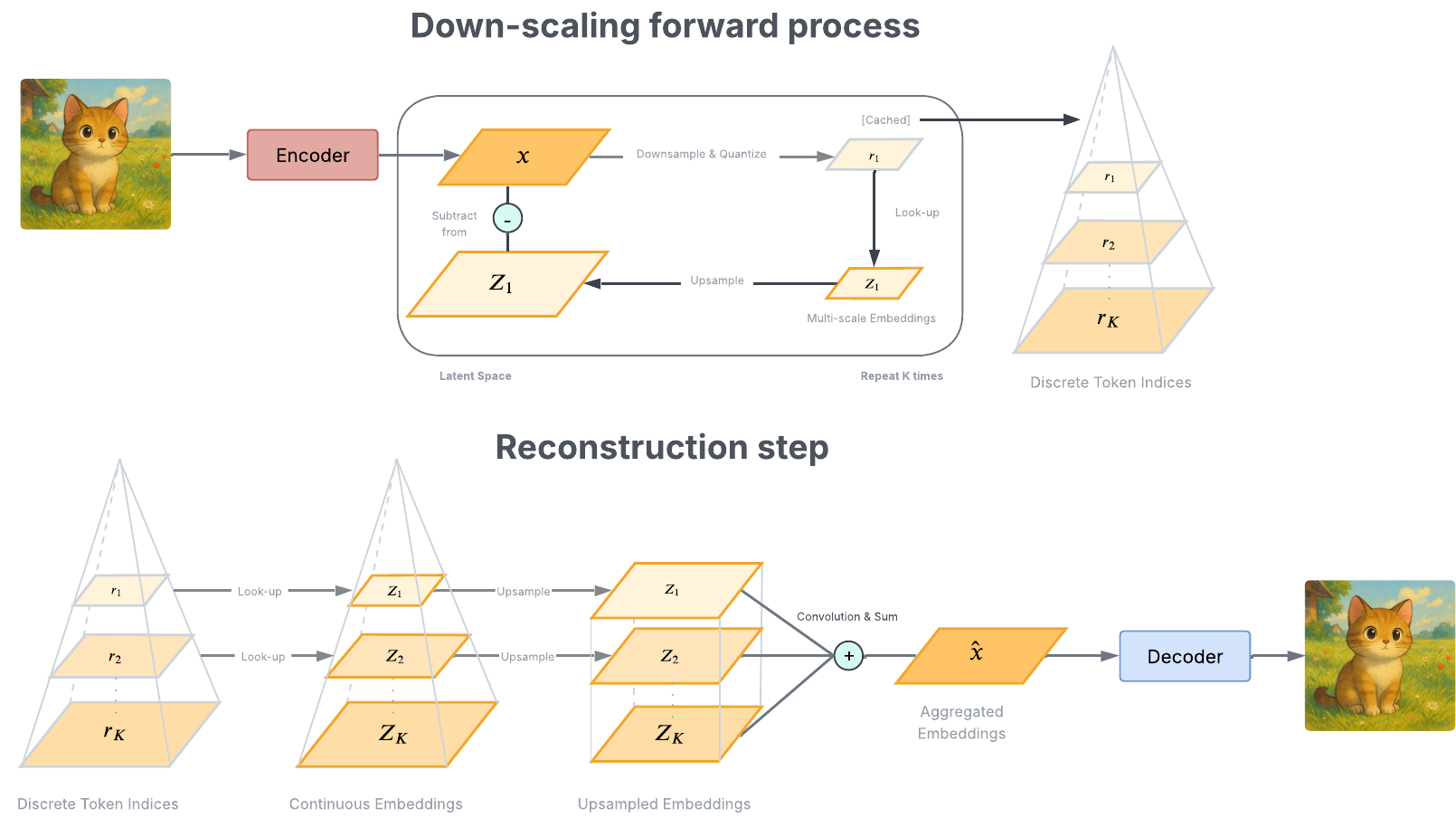}
  \caption{\textbf{Deterministic forward process in VAR.} An input image is encoded into a latent grid, then repeatedly downsampled to form a multi-scale pyramid. At each scale, the upsampled coarser latent is subtracted to obtain a residual, which is quantised into discrete code indices; the coarsest latent is quantised as base codes. The resulting set of base and residual codes defines deterministic training targets for the backward (generative) process.}
  \label{fig:forward}
\end{figure*}

\subsection{Rewriting VAR with iterative refinement}

We now define VAR as an iterative refinement framework. We work in the discrete latent space produced by a VQ-VAE tokeniser. The encoder maps $x$ to a continuous feature $z_e$, which is vector-quantised to an index grid $z^{(0)}$, and the decoder maps quantised latents back to pixels.

Let $L^{(0)}\in\mathbb{R}^{H_0\times W_0\times d}$ denote the $d$-channel latent obtained by embedding each index into $\mathbb{R}^d$. We construct a multi-scale pyramid by deterministic downscaling, $L^{(k)}=\mathcal{S}_k\!\big(L^{(k-1)}\big)$. In Laplacian form, the hierarchy is represented by residuals $R^{(k-1)}=L^{(k-1)}-\mathcal{U}_k\!\big(L^{(k)}\big)$. Each residual is quantised with a per-scale codebook in $\mathbb{R}^d$, and the coarsest latent $L^{(S)}$ is quantised as the base state.

\paragraph{The forward process.}
As in Figure \ref{fig:forward} first encode the image into latent space,
\begin{equation}
L^{(0)} \;=\; F(x).
\end{equation}
We then form a latent Gaussian pyramid by deterministic downscaling,
\begin{equation}
L^{(k)} \;=\; \mathcal{S}_k\!\bigl(L^{(k-1)}\bigr), \qquad k=1,\dots,S,
\end{equation}
and define latent residuals in Laplacian form,
\begin{equation}
R^{(k-1)} \;=\; L^{(k-1)} \;-\; \mathcal{U}_k\!\bigl(L^{(k)}\bigr), \qquad k=1,\dots,S.
\end{equation}
Each residual is pointwise quantised with a $d$-channel codebook, while the coarsest latent may be quantised as the base:
\begin{equation}
r^{(k-1)} \;=\; Q^{\mathrm{res}}_k\!\bigl(R^{(k-1)}\bigr), \quad k=1,\dots,S, 
\qquad
b^{(S)} \;=\; Q^{\mathrm{base}}_S\!\bigl(L^{(S)}\bigr).
\end{equation}
The forward distribution is therefore fully deterministic,
\begin{equation}
q\!\left(b^{(S)}, r^{(0)},\dots,r^{(S-1)} \mid x\right)
\;=\;
\delta\!\bigl(b^{(S)}-Q^{\mathrm{base}}_S(L^{(S)})\bigr)\;
\prod_{k=1}^{S}\,
\delta\!\bigl(r^{(k-1)}-Q^{\mathrm{res}}_k(R^{(k-1)})\bigr),
\end{equation}

\paragraph{The backward process.}
Generation proceeds from coarse to fine in latent space. At each scale, the model predicts residual indices in parallel and reconstructs latents via upsample-and-add.

Let the $d$-channel embeddings associated with the residual and base codebooks be $\mathrm{emb}^{\mathrm{res}}_k:\{0,\dots,V_k-1\}\to\mathbb{R}^d$ and $\mathrm{emb}^{\mathrm{base}}_S:\{0,\dots,V_S-1\}\to\mathbb{R}^d$. We first sample a factorised prior over the coarsest base codes and embed them,
\begin{equation}
b^{(S)} \sim p_\theta\!\bigl(b^{(S)}\bigr), 
\qquad
L^{(S)} \;=\; \mathrm{emb}^{\mathrm{base}}_S\!\bigl(b^{(S)}\bigr).
\end{equation}
For $k=S,\dots,1$ we form the upscaled context and a Transformer $\psi_\theta$ predicts the entire residual index map at the next finer scale in one shot,
\begin{equation}
C^{(k-1)} \;=\; \mathcal{U}_k\!\bigl(L^{(k)}\bigr), 
\qquad
p_\theta\!\bigl(r^{(k-1)} \mid L^{(k)}\bigr)
\;=\;
\prod_{i=1}^{H_{k-1}W_{k-1}}
\mathrm{Cat}\!\Bigl(r^{(k-1)}_i \,;\, \psi_\theta\!\bigl(C^{(k-1)}\bigr)_{i,\cdot}\Bigr),
\end{equation}
followed by residual embedding and latent reconstruction,
\begin{equation}
\widehat{R}^{(k-1)} \;=\; \mathrm{emb}^{\mathrm{res}}_k\!\bigl(r^{(k-1)}\bigr),
\qquad
L^{(k-1)} \;=\; C^{(k-1)} \;+\; \widehat{R}^{(k-1)}.
\end{equation}
The finest latent is finally decoded to pixels,
\begin{equation}
\hat{x} \;=\; G\!\bigl(L^{(0)}\bigr).
\end{equation}

\paragraph{Remark.}
In this formulation, \textbf{(i)} tokens within each scale are generated fully in parallel: conditioned on the upsampled context $C^{(k-1)}$, categorical factors across spatial sites are independent, so the entire residual map $r^{(k-1)}$ is produced in one shot.

\textbf{(ii)} The number of refinement steps is small: just $S$ scales (typically $S{=}8$ in the original setup), versus hundreds to a thousand time steps in common diffusion samplers.

\textbf{(iii)} The latent likelihood is exact and tractable because the model defines a finite autoregressive factorisation across scales. By contrast, DDPMs posit a first-order Markov chain $p_\theta(x_{0{:}T})=p(x_T)\prod_{t=1}^T p_\theta(x_{t-1}\!\mid x_t)$ and are trained/evaluated via an ELBO built from Gaussian KL terms.

\subsection{Discussion}
\label{sec:discussion}

Given this formulation, DDPM and VAR can both be viewed as iterative refinement methods. Under this shared view, three modeling choices are likely to matter for VAR's performance relative to DDPM.

\paragraph{Latent space.}
VAR refines in a learned latent representation rather than raw pixels. This typically lowers dimensionality and removes some local redundancy, so each update is applied to more compact, semantically organised features. A simpler target space can improve optimisation efficiency and may increase the effective signal-to-noise ratio during training, consistent with observations in latent diffusion \citep{rombach2022high}.

\paragraph{Discrete domain.}
VAR predicts categorical code indices, whereas standard DDPM predicts continuous noise. Casting prediction as classification can yield more stable gradients than direct regression in some settings \citep{van2017neural,razavi2019generating}. The finite codebook also imposes a structured representation that can support stepwise correction across scales. This does not imply that discrete targets are universally better; continuous noise prediction also has advantages, especially for smooth interpolation and flexible likelihood objectives \citep{ho2020denoising}.

\paragraph{Frequency refinement.}
VAR's next-scale updates implement an explicit coarse-to-fine decomposition, similar in spirit to a Laplacian pyramid \citep{burt1983multiresolution}. Early stages primarily model low-frequency structure, while later stages focus on high-frequency detail. This separation can reduce cross-frequency interference compared with updates that mix multiple frequency regimes in a single step. It also enables parallel token prediction within each scale. In practice, gains depend on the number of scales $S$, which remains a tunable design choice.

\section{Experiments}
\label{sec:mnist-exps}

We run three controlled MNIST \citep{deng2012mnist} studies to probe the hypotheses from Section \ref{sec:discussion}: latent-space refinement, discrete targets, and coarse-to-fine frequency decomposition. Results are reported as mean $\pm$ standard deviation over three seeds.

\begin{table}[h]
\centering
\caption{Common training protocol across experiments.}
\label{tab:common-setup}
\begin{tabular}{ll}
\toprule
Dataset & MNIST ($28\times28$, normalised to $[0,1]$) \\
Train/test corruption & Additive Gaussian noise, $\sigma=0.20$ unless stated \\
Backbone & 2-layer MLP, width 512, GELU, pre-output LayerNorm \\
Optimiser & AdamW, learning rate $10^{-3}$, weight decay $10^{-4}$ \\
Training budget & 30 epochs, batch size 256 \\
Latent encoder (when used) & Linear autoencoder, bottleneck $d=64$, then frozen \\
Pyramid filter (when used) & Separable blur $[1,4,6,4,1]/16$ \\
\bottomrule
\end{tabular}
\end{table}

\subsection{Latent vs Pixel Space}
We compare two denoisers that predict clean targets from noisy inputs $x+\varepsilon$. Pixel-MLP predicts $\hat{x}$ directly with an MSE objective in pixel space. Latent-MLP predicts $\hat{z}$ in the frozen latent space, supervised against $z=E(x)$, and reconstructs via $\hat{x}=D(\hat{z})$.

We evaluate test MSE, PSNR, epochs-to-target (MSE $\le 0.018$), and robustness under a test-time noise shift to $\sigma=0.30$.

\begin{table}[h]
\centering
\caption{Setup A: Pixel vs. latent denoising on MNIST.}
\label{tab:latent-vs-pixel}
\begin{tabular}{lcccc}
\toprule
Method & MSE @ $\sigma{=}0.20$ $\downarrow$ & PSNR (dB) $\uparrow$ & Epochs to $0.018$ $\downarrow$ & MSE @ $\sigma{=}0.30$ $\downarrow$\\
\midrule
Pixel-MLP  & $0.0195{\pm}0.0003$ & $17.6{\pm}0.1$ & $14{\pm}1$ & $0.0280{\pm}0.0010$ \\
Latent-MLP & $\mathbf{0.0158}{\pm}0.0002$ & $\mathbf{18.9}{\pm}0.1$ & $\mathbf{9}{\pm}1$  & $\mathbf{0.0221}{\pm}0.0010$ \\
\bottomrule
\end{tabular}
\end{table}

Latent-MLP reduces MSE from $0.0195$ to $0.0158$, improves PSNR from $17.6$ to $18.9$ dB, reaches the target in 9 epochs instead of 14, and remains better under the $\sigma=0.30$ shift. This is consistent with the hypothesis that compact learned representations improve optimization efficiency.

\subsection{Discrete vs Continuous Targets}
We build a $k$-means codebook ($k=64$) on clean latent features $E(x)$, then train a shared MLP trunk with alternative prediction heads. The regression head predicts continuous latent vectors with MSE. Softmax-64 predicts nearest-code indices with cross-entropy and dequantises to centroids. Bitwise-8 predicts a learned 8-bit code with independent Bernoulli heads.

We report dequantised test MSE, mini-batch gradient-norm variance (lower is steadier), and relative training time per epoch.

\begin{table}[h]
\centering
\caption{Setup B: Discrete prediction stabilises training at fixed capacity.}
\label{tab:disc-vs-reg}
\begin{tabular}{lccc}
\toprule
Head & Dequantised MSE $\downarrow$ & Grad var ($\times 10^{-3}$) $\downarrow$ & Time/epoch (rel.) \\
\midrule
Regression (MSE) & $0.0204{\pm}0.0004$ & $2.4{\pm}0.2$ & $1.00$ \\
Softmax-64 (CE)  & $\mathbf{0.0181}{\pm}0.0003$ & $1.5{\pm}0.1$ & $1.02$ \\
Bitwise-8 (BCE)  & $0.0185{\pm}0.0003$ & $\mathbf{1.1}{\pm}0.1$ & $\mathbf{0.92}$ \\
\bottomrule
\end{tabular}
\end{table}

Discrete targets improve both stability and accuracy at fixed capacity. Softmax-64 lowers MSE from $0.0204$ to $0.0181$ with lower gradient variance. Bitwise-8 further reduces gradient variance and is the fastest per epoch, indicating that factorised discrete targets can improve the quality-cost trade-off.

\subsection{Single-Stage vs Coarse-to-Fine Refinement}
We construct a two-level Laplacian decomposition with low-pass component $G_1$ ($14\times14$) and residual $R_0$ ($28\times28$). The single-stage baseline predicts the clean image directly from noisy input. The coarse-to-fine variant first denoises $G_1$ with $\mathrm{MLP}_{\text{coarse}}$, then predicts $R_0$ with $\mathrm{MLP}_{\text{resid}}$ conditioned on the upsampled $\widehat{G}_1$ and the noisy input, and reconstructs $\hat{x}=\mathrm{up}(\widehat{G}_1)+\widehat{R}_0$. Total parameters are matched to the baseline.

We evaluate test MSE, epochs-to-target (MSE $\le 0.018$), and high-frequency PSNR (HF-PSNR) computed on $R_0$.

\begin{table}[h]
\centering
\caption{Setup C: Coarse-to-fine specialisation reduces interference.}
\label{tab:coarse-fine}
\begin{tabular}{lccc}
\toprule
Method & MSE $\downarrow$ & Epochs to $0.018$ $\downarrow$ & HF-PSNR (dB) $\uparrow$ \\
\midrule
Single-shot (one MLP)       & $0.0193{\pm}0.0003$ & $12{\pm}1$ & $16.8{\pm}0.2$ \\
Coarse$\rightarrow$fine (2 MLPs) & $\mathbf{0.0164}{\pm}0.0002$ & $\mathbf{7}{\pm}1$  & $\mathbf{18.2}{\pm}0.2$ \\
\bottomrule
\end{tabular}
\end{table}

Increasing the number of bands beyond two yields diminishing returns:

\begin{table}[h]
\centering
\caption{Effect of refinement depth $S$ with matched total parameters.}
\label{tab:depth-sweep}
\begin{tabular}{lcc}
\toprule
$S$ (bands) & MSE $\downarrow$ & Train time/epoch (rel.) \\
\midrule
1 & $0.0193$ & $1.00$ \\
2               & $\mathbf{0.0164}$ & $1.15$ \\
3               & $0.0162$ & $1.22$ \\
\bottomrule
\end{tabular}
\end{table}

The coarse-to-fine model reduces MSE from $0.0193$ to $0.0164$, reaches the target 5 epochs earlier, and improves HF-PSNR by $1.4$ dB. The depth sweep shows that most gains appear at two bands, with only marginal improvement at three bands and higher per-epoch cost.

\section{Beyond Images: Applications of Iterative Refinement}

\subsection{Graph generation}
Graph generation \citep{you2018graphrnn, vignac2023digress} seeks to learn distributions over graphs for applications in chemistry, biology, and network science, where the target law should be invariant to node permutations.

A central challenge is ensuring permutation invariance of the induced distribution. Many autoregressive constructions impose an arbitrary node ordering, which can cause the likelihood or sampling trajectory to depend on that ordering and therefore break invariance \citep{zhao2024pard}. Typical remedies include using exchangeable priors and permutation-equivariant architectures (e.g., GNNs where message passing and pooling commute with node permutations), but care is still required in training objectives and decoding to avoid reintroducing order dependence.

Recent work \citep{belkadi2025diffusion} has instantiated a VAR-style, coarse-to-fine generator for graphs and reported competitive sample quality, along with significant inference speedups, suggesting that next-scale prediction may transfer beyond images.

Diffusion-based graph models, on the other hand, establish that, if the score/denoising network is permutation equivariant, the resulting generative distribution is permutation invariant \citep{niu2020permutation}.

\paragraph{Why the iterative refinement perspective may help.}
The iterative-refinement view suggests a direct design hypothesis for VAR on graphs. VAR already generates through a short sequence of refinement steps, analogous to diffusion updates. If each refinement map is implemented by a permutation-equivariant architecture (for example, a set-structured Transformer without order-dependent positional encodings or an equivariant GNN), and the coarsest prior is exchangeable, then the resulting generation process is expected to preserve permutation symmetry. This is a design principle rather than a formal proof for the full model class. Empirical checks of invariance and likelihood consistency across node relabelings are still required.

\subsection{Medium-range weather forecasting}
Medium-range (3–15 day) forecasts inform public safety and energy planning, where distribution-aware evaluation matters. GenCast \citep{price2023gencast} established a skilful, probabilistic ML forecaster by framing the task as generative modelling with diffusion, providing calibrated ensemble trajectories rather than a single path. In contrast, leading deterministic ML forecasters (e.g., GraphCast \citep{Lam2023GraphCast}, Pangu \citep{Bi2023Pangu}) are single-pass generators trained for point predictions and do not provide distributional forecasts.

\paragraph{Why the iterative refinement perspective may help.}
An iterative-refinement VAR-style forecaster could be probabilistic by construction if each scale predicts distributions over discrete latents. In that case, multiple ensemble members can be sampled with low incremental cost per member. The coarse-to-fine decomposition also aligns with atmospheric scale separation: early levels represent planetary and synoptic structure, while later levels refine mesoscale detail.

\section{Limitations and Future Work}

\paragraph{Scale of experiments.}
Our empirical study is intentionally limited in scope. We use controlled MLP baselines to isolate specific factors under modest compute budgets. A broader assessment should train a full VAR model end-to-end on standard image benchmarks and compare it with a diffusion baseline under matched architectures and training budgets. Useful ablations include: (i) latent vs. pixel, by running next-scale refinement in VQ-VAE latent space and in a Laplacian pixel pyramid as a no-latent control; (ii) discrete vs. continuous, by replacing categorical prediction heads with Gaussian regression heads while keeping the same pyramid structure; and (iii) multi-scale vs. single-scale, by collapsing to a single latent grid. For each setting, one can add a diffusion objective in the same representation space (pixel, latent, or discrete) and compare quality--cost trade-offs (e.g., FID/PSNR versus wall-clock and function evaluations), calibration (e.g., bits/dim with standard dequantisation), and sensitivity to refinement depth $S$ versus diffusion steps $T$.

\paragraph{Deterministic forward process.}
In standard VAR, the forward construction of the latent/residual pyramid is deterministic. Stochasticity is introduced only in some variants, such as bitwise self-correction methods that inject Bernoulli noise at the code level \citep{han2025infinity}. The effect of adding controlled stochasticity to the forward process is not yet clear and warrants systematic evaluation.

\paragraph{Leveraging diffusion methods.}
The iterative-refinement view suggests several diffusion-inspired techniques that may be applicable to VAR. Candidate directions include classifier-free guidance for categorical heads, self-conditioning across scales, loss reweighting, few-step distillation from larger-$S$ models, and schedule design or early-exit strategies. Other possibilities include consistency-style objectives between scales, latent-consistency adapters for faster sampling, and per-scale temperature or top-$k$ calibration to tune fidelity-diversity trade-offs. These directions should be treated as hypotheses for future empirical study.

\section{Conclusion}
We reframed Visual Autoregressive models as iterative refiners operating on a latent Laplacian pyramid. This view makes the generation path explicit as a short sequence of coarse-to-fine residual updates and clarifies how VAR relates to denoising diffusion while retaining exact cross-scale factorisation and within-scale parallelism. Under this lens, three design choices appear central: working in a compact latent space, predicting discrete indices rather than continuous targets, and allocating supervision across frequency bands. Small MNIST toy experiments support these hypotheses, indicating that a modest number of banded steps can capture most of the benefit.

Beyond images, the same template suggests practical routes for permutation-equivariant graph generation and for efficient probabilistic weather forecasting, where calibrated ensembles are required. The formulation also exposes straightforward interfaces to the diffusion ecosystem, including guidance, consistency-style objectives, and few-step distillation adapted to categorical heads and scale schedules. A fuller assessment will require end-to-end studies with matched backbones on standard benchmarks, but the iterative-refinement framing already offers a common language for analysing, ablating, and improving both VAR and diffusion families.

\newpage
\bibliographystyle{unsrtnat}
\bibliography{references}

\appendix

\end{document}